# PPO-Based Dynamic Control of Uncertain Floating Platforms in Zero-G Environment

Mahya Ramezani, M. Amin Alandihallaj, and Andreas M. Hein

*Abstract*— In the field of space exploration, floating platforms play a crucial role in scientific investigations and technological advancements. However, controlling these platforms in zero-gravity environments presents unique challenges, including uncertainties and disturbances. This paper introduces an innovative approach that combines Proximal Policy Optimization (PPO) with Model Predictive Control (MPC) in the zero-gravity laboratory (Zero-G Lab) at the University of Luxembourg. This approach leverages PPO's reinforcement learning power and MPC's precision to navigate the complex control dynamics of floating platforms. Unlike traditional control methods, this PPO-MPC approach learns from MPC predictions, adapting to unmodeled dynamics and disturbances, resulting in a resilient control framework tailored to the zero-gravity environment. Simulations and experiments in the Zero-G Lab validate this approach, showcasing the adaptability of the PPO agent. This research opens new possibilities for controlling floating platforms in zero-gravity settings, promising advancements in space exploration.

## I. INTRODUCTION

The pursuit of space exploration rests on a foundation of meticulous testing and validation, a cornerstone that not only enhances the reliability of space missions but also augments their operational efficiency. The complexities inherent in the frictionless environment necessitate ground-based testing to mirror the conditions and challenges faced by spacecraft and satellites in orbit. To address this need, cutting-edge ground test facilities have emerged as indispensable tools in the arsenal of space research and development.

The Georgia Institute of Technology's ASTROS facility stands as a hub for spacecraft Autonomous Rendezvous and Docking (ARD) maneuvers, wielding high-pressure air-bearing floating platforms over a 4m x 4m flat epoxy floor to simulate frictionless operations [1]. The European Space Agency's ORBIT facility, spanning 45 m2 epoxy floor, excels in orbital robotics, leveraging air-bearing platforms for position tracking and facilitating large payload tests [2]. ADAMUS, a 6-DoFs spacecraft simulator at the forefront of autonomy research, graces the scene with torque and force-free operation [3]. The Spacecraft Dynamics Simulator at Caltech reveals a multifaceted multi-Spacecraft testbed, featuring M-STAR platforms for 3 to 6-DoFs experiments [4]. AUDASS, a standout from the Satellite Servicing Laboratory, embodies independent floating platforms via air-bearings, an embodiment of proximity maneuvers [5]. NASA's contributions encompass the Air Bearing Floor at Johnson Space Center and the Formation Control testbed at JFP, highlighting suspended platforms and precision formation flight, respectively [6]. This global panorama of facilities collectively propels our understanding of space dynamics, steering the evolution of control strategies for the uncharted frontiers of space exploration.

Among these test facilities, the Zero-G Lab [7] at the University of Luxembourg, shown in Fig. 1, stands as a pioneering exemplar. Within this controlled environment, researchers and engineers are afforded the opportunity to scrutinize the performance of space technologies and systems in space conditions. Central to this endeavor is the utilization of a sophisticated mechatronic system, the floating platform, engineered to simulate the complexities of space operations.

The floating platform serves as a conduit to assess diverse scenarios of space missions, encompassing rendezvous and docking maneuvers, relative motion of satellites, and intricate orbital scenarios. By scrutinizing these scenarios, the Zero-G Lab contributes not only to our understanding of space dynamics but also to the refinement of control strategies vital for the success of space missions.

M. Ramezani is with the Automation and Robotics Research Group, Interdisciplinary Centre for Security, Reliability and Trust (SnT), University of Luxembourg (UL), Luxembourg (corresponding author; e-mail: mahya.ramezani@uni.lu).

M. A. Alandihallaj is with the Space Systems Research Group, Interdisciplinary Centre for Security, Reliability and Trust (SnT), University of Luxembourg (UL), Luxembourg (e-mail: amin.hallaj@uni.lu).

A. M. Hein is with the Space Systems Research Group, Interdisciplinary Centre for Security, Reliability and Trust (SnT), University of Luxembourg (UL), Luxembourg (e-mail: andreas.hein@uni.lu).

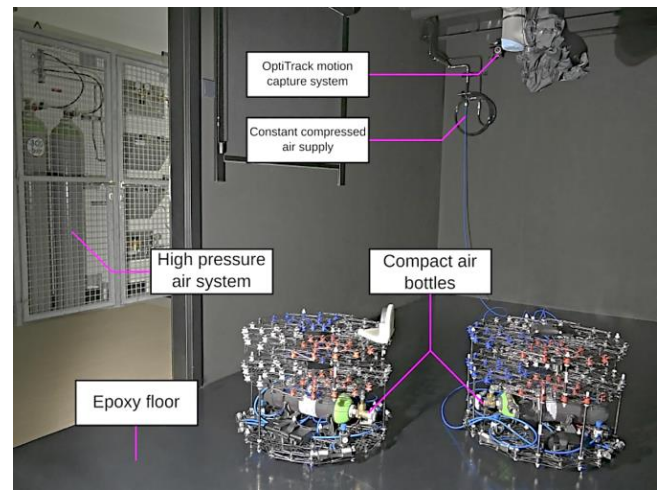

Figure 1 The major components of the Zero-G Lab [8].

However, achieving effective control of floating platforms is a challenging endeavor. Conventional control methods struggle to handle the complexities inherent in controlling the coupled dynamics of these frictionless floating platforms [9]. Uncertainties, unmodeled dynamics, and external disturbances, worsened by the inadvertent incline of the lab floor, collectively conspire to undermine the stability and precision of floating platform control [10, 11].

Despite the unique features and versatility of Model Predictive Control (MPC), which make it suitable for various space applications such as space tether control [12], satellite formation flight control [13, 14], spacecraft rendezvous control [15, 16], satellite attitude control [17], satellite maneuvering planning [18], and asteroid landing control [19, 20] and hovering [21], its application in stabilization of floating platforms has not yielded the desired performance.

To surmount these challenges, this paper introduces a transformative approach anchored in the Proximal Policy Optimization (PPO) method. Given the inherent unpredictability and incomplete knowledge of the environment, the use of machine learning methods emerges as a potential solution [22]. This paper introduces and presents the outcomes of adopting a transformative approach centered around the PPO method for stabilizing the floating platform. This approach has demonstrated satisfactory performance levels in controlling space systems [23, 24]. To enhance learning efficiency [25], PPO is integrated with MPC, creating a novel paradigm that leverages Reinforcement Learning (RL) to navigate dynamic and uncertain control scenarios. By capitalizing on learning from experience, the PPO-based approach offers a pathway to transcend the limitations of traditional control methods, presenting adaptability and resilience in the face of the system's unique challenges.

The primary aim of this paper is to present and evaluate the efficacy of the PPO-based control approach within the context of the Zero-G Lab at the University of Luxembourg. The paper outlines the foundational principles of PPO, elucidates its integration with MPC for enhanced learning, describes the experimental setup, and analyzes the results of both simulations and empirical trials. Through this inquiry, the paper seeks to validate the potential of PPO in conquering the complexities of zero gravity control and to contribute to the advancement of adaptable control strategies tailored for space environments.

## II. METHODOLOGY

RL is a crucial branch of machine learning dedicated to optimizing policies that link observations to actions, which aims to maximize rewards accumulated through trajectories in an environment, as agents adjust actions based on rewards [26]. This involves a Markov decision process defined by state space, action space, state transitions, and rewards.

Based on [27], trajectories, $\tau$, are core units in RL, representing sequences of state-action pairs during episodes. Mathematically, $\tau = [x_0, u_0, ..., x_T, u_T] \in T$, with $x$ as state, $u$ as action, and $T$ as steps. The main RL goal is to optimize the expectation of cumulative rewards across trajectories as

$$\mathrm{E}_{p_\alpha(\tau)}[r(\tau)] = \int_T r(\tau) p_\alpha(\tau) d\tau \tag{1}$$

where $r(\tau) = \sum_{t=0}^{T} \gamma^t r(x_t, u_t)$ is the summation of discounted rewards using $\gamma \in (0,1)$, $p_\alpha(\tau) = \prod_{t=0}^{T-1} p(x_{t+1} \mid x_t, u_t) \cdot p(x_0)$ is the probability of a trajectory under $\alpha$, and $u_t$ is a sample from $\pi_\alpha(u_t \mid x_t)$. The policy's conditional distribution introduces stochasticity in action choice, aiding exploration. As learning progresses, variance diminishes, favoring policy exploitation. Post-learning, policy variance becomes zero, ensuring deterministic action selection. This transition to determinism governs practical implementation.

The policy (actor) and the advantage function (critic) evolve simultaneously in PPO. PPO utilizes the state-value function $V_w^\pi(x_t) = \mathrm{E}_\pi(\sum_{k=t}^{T} \gamma^{k-t} r_k(x_k, u_k) \mid x_t)$ to estimate discounted rewards across trajectories. The parameter vector $w$ and the policy parameter vector $\alpha$ are learned during the learning process. The resultant advantage function $A_w^\pi(x_t, u_t)$ quantifies the difference between empirical and estimated rewards.

$$A_w^\pi(x_t, u_t) = \left(\sum_{k=t}^{T} \gamma^{k-t} r_k(x_k, u_k)\right) - V_w^\pi(x_t) \tag{2}$$

PPO, a successor of the trust region policy optimization algorithm, retains the ability to mitigate substantial policy updates, reducing the risk of learning divergence. This while maintaining a simpler and more widely implementable approach. At the core of PPO lies the policy probability ratio $p_t(\alpha) = \frac{\pi_\alpha(u_t \mid x_t)}{\hat{\pi}_\alpha(u_t \mid x_t)}$, which gauges the probability of selecting an action after a learning update, $\pi_\alpha(u_t \mid x_t)$, compared to before the update, $\hat{\pi}_\alpha(u_t \mid x_t)$. This ratio directly informs the PPO loss function as follows.

$$\mathcal{L}(\alpha) = \mathrm{E}_{p(\tau)}\left[\min\left(\begin{array}{c} p_t(\alpha) A_w^\pi(x_t, u_t), \\ \mathrm{clip}[p_t(\alpha), \epsilon] A_w^\pi(x_t, u_t) \end{array}\right)\right] \tag{3}$$

Here, the clip function, given by

$$\mathrm{clip}[p_t(\alpha), \epsilon] = \begin{cases} 1 - \epsilon & p_t(\alpha) < 1 - \epsilon \\ 1 + \epsilon & p_t(\alpha) < 1 + \epsilon \\ p_t(\alpha) & \text{otherwise} \end{cases} \tag{4}$$

imposes bounds on the policy probability ratio using a clipping parameter $\epsilon$ within $(0,1)$. This constrains policy updates, facilitating a trust region to eliminate unwarranted changes. Notably, the loss function is relative to the policy pre-update, making its absolute value across multiple updates less informative. Instead, its immediate gradient plays a pivotal role in steering the policy to optimize rewards over all trajectories. To learn the state-value function, a commonly utilized mean squared error cost function is minimized

$$L(w) = \frac{1}{2} \mathrm{E}_{p(\tau)}\left[\left(V_w^\pi(x_t) - \left[\sum_{k=t}^{T} \gamma^{k-t} r(x_t, u_t)\right]\right)^2\right] \tag{5}$$

with an objective function for policy enhancement and a cost function for state-value correction, gradients of these

functions facilitate gradient ascent on $\boldsymbol{\alpha}$ and gradient descent on $\boldsymbol{w}$

$$\begin{aligned}\boldsymbol{\alpha}_+ &= \boldsymbol{\alpha}_- + \beta_\alpha \nabla_\alpha J(\boldsymbol{\alpha})|_{\alpha=\alpha_-} \\ \boldsymbol{w}_+ &= \boldsymbol{w}_- - \beta_w \nabla_w L(\boldsymbol{w})|_{w=w_-}\end{aligned} \quad (6)$$

Here, $\beta_\alpha$ and $\beta_w$ are learning rates for policy and state-value function respectively, set by the designer.

To develop policies for 3-DOF maneuvers within the lab, we adopt the lab-centered inertial frame I. The state vector $\boldsymbol{s} = [\boldsymbol{r}, \boldsymbol{v}, \boldsymbol{\theta}, \boldsymbol{\omega}]$ represents the floating platform's center of mass within the I frame, with $\boldsymbol{r} \in \mathbb{R}^2$ as position, $\boldsymbol{v} \in \mathbb{R}^2$ as velocity, $\boldsymbol{\theta} \in \mathbb{R}^1$ as attitude angle, and $\boldsymbol{\omega} \in \mathbb{R}^1$ as angular velocity. The control action $\boldsymbol{u} = [\boldsymbol{F}, \boldsymbol{M}]$ includes thrust command $\boldsymbol{F} \in \mathbb{R}^2$ and torque command $\boldsymbol{M} \in \mathbb{R}^1$, both in the floating platform body frame, $\mathcal{B}$, subject to actuator constraints. Dynamics are derived in continuous-time and discretized with a 0.1-second sample period. Translational dynamics are modeled as double integrators:

$$\begin{aligned}\dot{\boldsymbol{r}} &= \boldsymbol{v} \\ \dot{\boldsymbol{v}} &= \frac{C_I^\mathcal{B}(\boldsymbol{\theta})\boldsymbol{F}_\mathcal{B}}{m}\end{aligned} \quad (7)$$

where $m$ is the floating platform mass and $C_I^\mathcal{B}(\boldsymbol{\theta}) \in \mathbb{R}^{2\times 2}$ represents the rotation matrix that maps from the $\mathcal{B}$ to the I frame.

Attitude dynamics follow quaternion kinematics and Euler's equations for a rigid body:

$$\begin{aligned}\dot{\boldsymbol{\alpha}} &= \boldsymbol{\omega} \\ \dot{\boldsymbol{\omega}} &= \frac{\boldsymbol{L}}{J}\end{aligned} \quad (8)$$

in which $J$ is the moment of inertia of the floating platform around the rotation axis.

Policy and state-value functions are modeled with feedforward neural networks, parameterized by $\boldsymbol{\alpha}$ and $\boldsymbol{w}$, which are updated based on (6) using Adam optimizer [28]. The policy is a multivariate Gaussian distribution with a diagonal covariance matrix. Neural network outputs are scaled using running mean and standard deviation of experienced state data during learning. Policy network outputs are scaled so that $\pm 1$ corresponds to maximum/minimum thrust or torque.

The actor network is structured as a feed-forward neural network consisting of two hidden layers, each comprising [128, 64] neurons. Meanwhile, the critic network exhibits a more intricate architecture with three layers housing [128, 64, 8] neurons. Both networks utilize the tanh activation function. The output layer of the actor network comprises 3 neurons with linear activation, while the critic function incorporates one neuron with linear activation.

In the PPO implementation, a strategy inspired by Gaudet et al. [29] is employed, whereby learning parameters are dynamically adapted to achieve a desired target Kullback-Leibler (KL) divergence value between successive policy updates [30]. This approach is utilized to prevent significant policy updates that could potentially disrupt the learning process, ensuring that policy updates proceed gradually and with stability. Throughout the learning process, both $\epsilon$ and $\beta_\alpha$ are continuously adjusted to ensure that the KL-divergence between updates remains as close as possible to the specified target value $KL_d$.

Ensuring a well-defined reward function is paramount to the efficacy of PPO, as the policy's learning process centers on maximizing this function. In the context of 3-DOF stabilization maneuvers, the reward function encompasses multiple components, collectively addressing the minimization of state tracking discrepancies, control input exertion, and the reinforcement of successful stabilization outcomes. These components are deliberately assigned relative weights through design coefficients.

The primary term serves as a crucial component in aiding PPO's learning process from MPC. It quantifies the quadratic-weighted difference between the state derivatives produced by the RL agent and a reference obtained from MPC. This inclusion accelerates the learning process by providing a clear reward signal across the entire state-space, guiding the RL agent toward the attainment of effective stabilizing trajectories.

MPC involves minimizing a cost function that measures the difference between the floating platform's current and desired final stabilization states, along with control inputs, while considering dynamics and constraints. MPC iteratively solves an optimization problem, adapting control inputs in real-time to address uncertainties and disturbances like fuel sloshing.

The optimization problem of MPC is expressed as follows:

$$\begin{aligned}&\text{Minimize}_{\boldsymbol{u}(t)} \\ &\int_{t=t_0}^{t_f}\left[\|\boldsymbol{s}'(t) - \boldsymbol{s}_d(t)\|_\Omega^2 + \|\boldsymbol{u}(t)\|_\rho^2\right]dt \\ &\text{Subject to:} \\ &\quad \dot{\boldsymbol{s}}'(t) = \widehat{\boldsymbol{A}}\boldsymbol{s}'(t) + \widehat{\boldsymbol{B}}\boldsymbol{u}(t) \\ &\quad \boldsymbol{s}'(t) \in \Sigma \\ &\quad \boldsymbol{u}(t) \in \mathbf{U} \\ &\quad \boldsymbol{s}'(t_0) = \boldsymbol{s}_t\end{aligned} \quad (9)$$

where $t_f$ represents the final stabilization time, $t_0$ represents the current time instant, $\boldsymbol{s}'(t)$ represents the state vector of the linearized system, $\boldsymbol{s}_d$ represents the desired states, $\|.\|_\Omega^2$ denotes the weighted norm of a quantity defined by $(.)^T \Omega (.)$, with $\Omega$ being a positive definite matrix, $\boldsymbol{u}(t)$ represents the control input, $\widehat{\boldsymbol{A}}$ and $\widehat{\boldsymbol{B}}$ are system matrices representing the linearized dynamics of the floating platform, can be found in [9], $\Sigma$ is the set of feasible states representing constraints, $\mathbf{U}$ is the set of feasible control inputs representing constraints.

The MPC approach with a prediction horizon of 10s and a time step of 0.1s is utilized to generate the reference trajectory. Furthermore, $\Omega$ is represented as a diagonal matrix with elements set to 1 for position and angle-related diagonal elements, and 100 for time derivative-related elements. Additionally, $\boldsymbol{\rho}$ corresponds to a diagonal matrix with all diagonal elements equal to 1000.

The reward function is defined as follows.

$$r(\boldsymbol{x_t}, \boldsymbol{u_t}) = -\underbrace{\|\dot{\boldsymbol{s}}'_t - \dot{\boldsymbol{s}}_t\|_M^2}_{\Psi_1} - \underbrace{\|\boldsymbol{u}_t\|_P^2}_{\Psi_2} + \frac{\Psi_1}{1+e^{k(t)\delta}} + \frac{\Psi_2}{1+e^{k(t)\sigma}} \quad (10)$$

where the first term serves to establish the quadratic weighted error between the output generated by the RL agent and MPC, while the second term represents the quadratic control cost aimed at minimizing control effort. The third term corresponds to the terminal stabilization bonus. Here, $k(t) > 0$ denotes a monotonically increasing function, $\sigma$ and $\delta$ are the position and rotation angle stabilization errors, respectively, and $\Psi_{1,2} > 0$. The terminal stabilization bonus operates such that the reward increases exponentially as the system approaches the stabilization state. Additionally, as the parameter $k$ is increased, the reward range becomes progressively narrower and smaller.

It should be noted that to implement control commands, a Pulse-Width Pulse-Frequency (PWPF) modulator is employed. This modulation technique converts continuous analog control commands into discrete on/off signals for the thrusters. The PWPF modulator incorporates a Schmidt trigger and a first-order filter, adjusting the width and frequency of control pulses to regulate the thrust amplitude efficiently. This modulation technique offers advantages, such as reduced fuel consumption and improved accuracy compared to classical on/off controllers [31].

During the training episode, termination occurs either upon satisfaction of the stabilization requirements or when the time limit is reached.

## III. EXPERIMENTAL SETUP

The experimental environment is the Zero-G Lab, which features a spacious experimental room measuring 5m x 3m x 2.3m, equipped with two floating platforms that move frictionlessly over a meticulously installed epoxy floor. To maintain a near-frictionless environment, the floating platform is equipped with air-bearings that direct high-pressurized air towards the epoxy floor, eliminating mechanical contact [32]. The actuation of eight nozzles drives the floating platform along two translational axes, X and Y, as well as one rotational axis, Z ($\boldsymbol{\theta}$). These nozzles can generate forces up to 1N under pressures of 10 bar. Yalçın, et al. [9] provide comprehensive information regarding the order and locations of nozzles around the floating platform. Tracking the position of the floating platform is accomplished using six OptiTrack Prime 13W cameras located within the Zero-G Lab, operating at 240 Hz. An active marker positioned at the center of the top plate facilitates this tracking process [33]. The floating platform seamlessly integrates into the ROS network, and a ROS-MATLAB bridge facilitates platform programming using MATLAB, enabling experimentation and assessment of its capabilities. Fig. 2 illustrates how the system works.

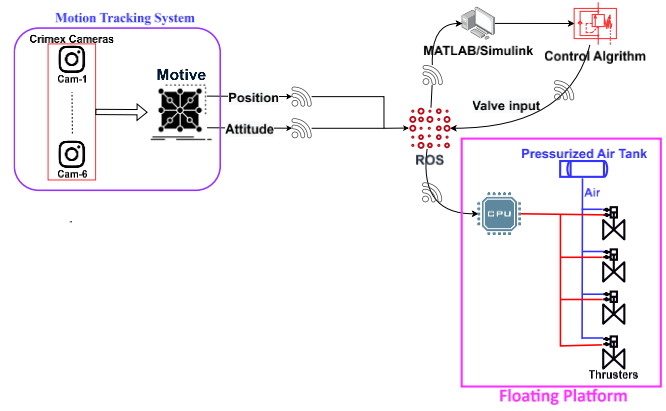

Figure 2. The system data flow in the Zero-G Lab.

## IV. TRAINING

The algorithm proposed in this study operates within the MATLAB environment, with a maximum iteration limit set at 20,000 to ensure network convergence. The agent collects data in batches of 200 episodes before performing policy and state-value function learning updates based on (6).

During both training and testing episodes, the policy generates force/torque commands at discrete 0.1s intervals. Training episodes have a time limit of 60s to efficiently gather data while ensuring stabilization. However, for testing, the time limit is extended to 100s to allow valid stabilization trajectories to complete. Furthermore, the acceptance stabilization condition requires an accuracy of 0.05m in distance, a velocity within the range of ±0.1 m/s, a rotation of up to ±5 degrees, and an angular velocity within the range of ±1 degree per second.

One objective of this research is to develop a robust feedback control law capable of handling significant uncertainty in the initial conditions of the stabilization maneuver. The RL goal is to create a stabilization policy effective across a wide range of initial conditions, encompassing the required robustness against uncertainty. Table I lists the training parameters.

The system's performance is evaluated in comparison to the PPO-only method. In the PPO-only method, the reference state time derivatives in the reward function (10) are set to zero.

TABLE I. THE TRAINING SETTING PARAMETERS.

| Parameter | Value |
|---|---|
| $KL_d$ | 0.001 |
| $\gamma$ | 0.98 |
| M | diag(1,10,5) |
| P | diag(10,10,10) |
| $\Psi_1$ | 100 |
| $\Psi_2$ | 10 |

As illustrated in Fig. 3, the graph depicts the normalized average cumulative reward over the training phase. Notably, it showcases that the integrated PPO with MPC outperforms the PPO-only approach. The integrated PPO-MPC exhibits a higher reward at the conclusion of the training phase, and its convergence rate is notably faster compared to the PPO-only method. This performance disparity can be attributed to the fact that in the PPO-MPC approach, PPO leverages optimal solutions learned from MPC, leading to quicker convergence toward an optimal behavior. In contrast, the PPO-only method necessitates an extensive search process and may become trapped in a local minimum that is inferior to the result obtained from MPC. However, it is plausible that the PPO-only method could eventually converge to the performance level of PPO-MPC, but it would require a more extended training phase. It is worth noting that, as evident from the graph, after 20,000 episodes, the average reward of the PPO-only approach has not yet converged, indicating the need for additional episodes to achieve convergence.

## V. EXPERIMENTAL RESULTS AND DISCUSSION

After training both the PPO-MPC and PPO-only methods with 20,000 episodes, their performance is assessed in real-world experiments involving the Floating Platform in the Zero-G Lab. During these experiments, the floating platform is manually disturbed four times at different intervals, and the objective is for it to autonomously return to the stabilization condition at the center of the lab. The times at which disturbances are introduced are highlighted with red arrows in the figures.

### A. Performance of PPO-MPC

Fig. 4 illustrates the performance of the PPO-MPC approach. Each time the platform is disturbed, it effectively returns to the stabilization condition, and the state errors remain within the predefined range (0.05m in distance and a rotation error of up to ±5 degrees). Notably, the second disturbance exhibits a more significant rotation angle deviation, while the other disturbances primarily affect the platform's position, with less impact on orientation. The actuation of the thrusters, as demonstrated in Fig. 5, showcases the successful generation of pulse signals for each individual nozzle using the PWPF method.

### B. Performance of PPO-Only

In contrast, Fig. 6 displays the performance of the PPO-only method. While the platform does return to its origin after disturbances, both the stabilization error and the return time exceed the predefined thresholds. The stabilization position error measures around 0.15m, and the rotation error is approximately 10 degrees. This outcome was anticipated, as the PPO-only method achieved lower rewards during the training phase compared to the PPO-MPC approach. Consequently, the PPO-MPC integration demonstrates superior performance, aligning with expectations. The nozzle actuation for the PPO-only case is depicted in Fig. 7.

The experimental results underscore the effectiveness of the integrated PPO-MPC method in achieving precise and rapid stabilization of the floating platform under disturbance, outperforming the PPO-only approach in this space environment.

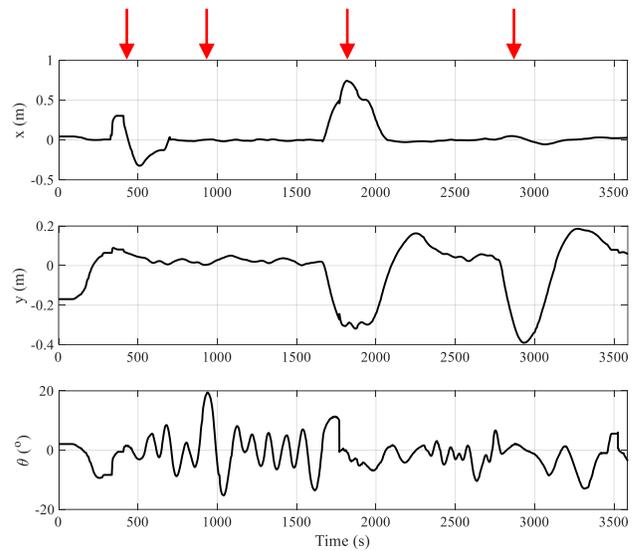

Figure 4. Performance of the PPO-MPC approach in disturbance rejection at Zero-G Lab experiment.

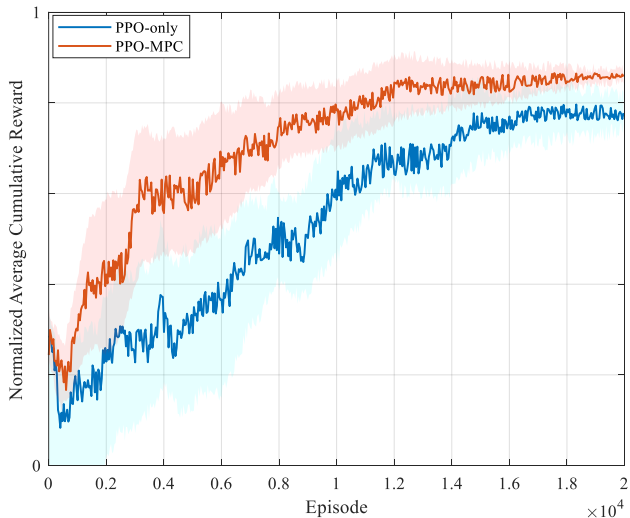

Figure 3. The normalized average cumulative reward in the training phase.

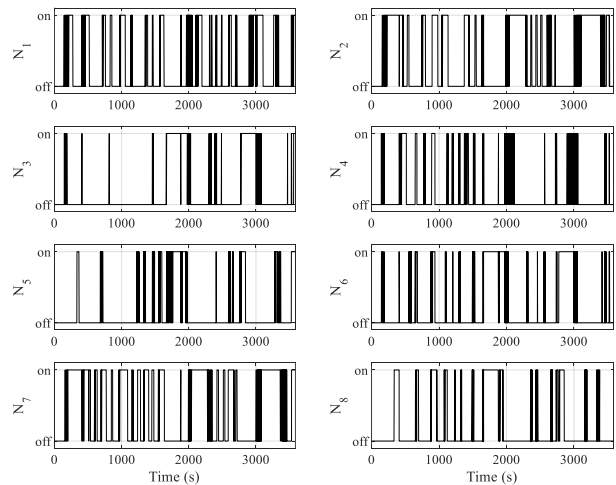

Figure 5. Thruster actuation in PPO-MPC control.

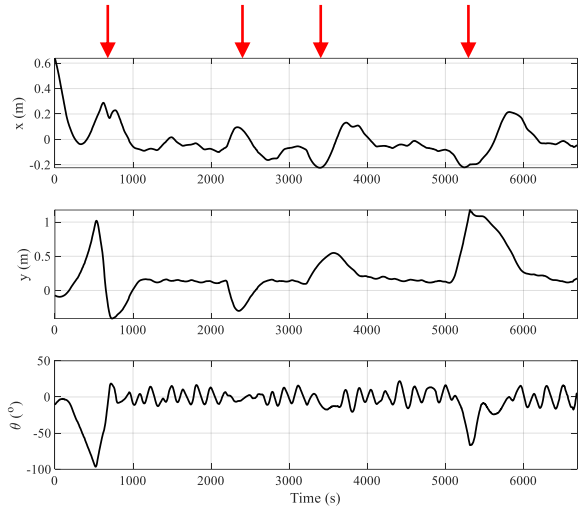

Figure 6. Performance of the PPO-only approach in disturbance rejection at Zero-G Lab experiment.

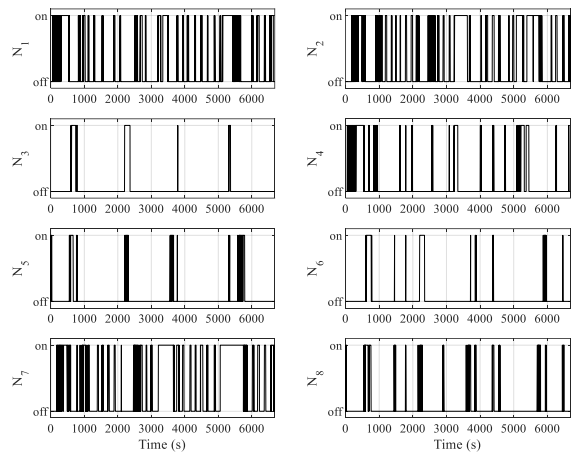

Figure 7. Thruster actuation in PPO-only control.

## VI. CONCLUSION

In this study, we have explored the use of Proximal Policy Optimization (PPO) combined with Model Predictive Control (MPC) for the control of a floating platform within the unique environment of the Zero-G Lab. Through extensive training and real-world experiments, we have gained valuable insights and drawn important lessons regarding the control of such platforms in a frictionless, zero-gravity setting.

Our research has yielded several key lessons that have significant implications for the control of floating platforms in space environments:

### Adaptability Through Integration

The integration of PPO with MPC has proven to be a powerful strategy. This combined approach leverages the predictive capabilities of MPC to enhance the adaptability of PPO. The result is a control framework that quickly responds to disturbances and converges to optimal solutions. This adaptability is crucial for effectively dealing with uncertainties and unmodeled dynamics inherent in space settings.

### Robustness is Paramount

The experiments conducted in the Zero-G Lab underscore the importance of robust control strategies. The PPO-MPC approach consistently outperformed the PPO-only method in terms of robustness and precision. It was able to counteract disturbances effectively, returning the platform to its desired state with minimal errors. This robustness is a critical factor for ensuring the success of missions in space exploration.

### Speed of Learning Matters

The PPO-MPC approach exhibited a faster convergence rate during training compared to the PPO-only method. This speed of learning is essential, especially in dynamic and uncertain environments. It allows the control system to adapt quickly to changing conditions, which is crucial for maintaining stability and achieving mission objectives.

### Implications for Space Exploration

The lessons learned from this study have significant implications for space exploration. Precise control of floating platforms is essential for various scientific investigations and technological advancements in space environments. The adaptability and robustness demonstrated by the PPO-MPC approach make it a promising candidate for addressing the control challenges encountered in space missions.

In summary, the integration of PPO with MPC has unveiled new horizons in the realm of controlling floating platforms in zero-gravity environments. The knowledge acquired from this study paves the way for more effective and reliable control strategies in the context of space exploration. In this domain, precision and adaptability are not mere aspirations but prerequisites for unraveling the mysteries of the cosmos and advancing our understanding of the universe.

ACKNOWLEDGMENT

We extend our sincere gratitude to Space Robotics (SpaceR) Research Group at the Interdisciplinary Centre for Security, Reliability, and Trust (SnT) of the University of Luxembourg, with special thanks to Dr. Baris Can Yalcin, for the invaluable collaboration in conducting the experiments at the Zero-G Lab. Their expertise and support have been instrumental in the successful execution of this research.